\documentclass[10pt]{article}

\usepackage[preprint]{tmlr}

\usepackage{amsmath,amsfonts,bm}

\def\eqref#1{equation~\ref{#1}}

\def\1{\bm{1}}

\DeclareMathAlphabet{\mathsfit}{\encodingdefault}{\sfdefault}{m}{sl}
\SetMathAlphabet{\mathsfit}{bold}{\encodingdefault}{\sfdefault}{bx}{n}

\newcommand{\boldW}{{\boldsymbol{W}}}

\newcommand{\boldv}{{\boldsymbol{v}}}

\usepackage{hyperref}
\usepackage{url}
\usepackage{graphicx, color}

\usepackage[noorphans]{quoting}
\usepackage[square,sort,comma,numbers]{natbib}

\usepackage{amsmath}
\usepackage{amssymb}
\usepackage{mathtools}
\usepackage{amsthm}
\usepackage{bbm}
\usepackage{booktabs}
\usepackage{tabularx}
\usepackage{subcaption}
\usepackage{wrapfig}

\usepackage{caption}

\usepackage[most]{tcolorbox}

\usepackage[ruled, algo2e, vlined]{algorithm2e}
\SetKwInput{KwInput}{Input}
\SetKwInput{KwOutput}{Output}

\theoremstyle{plain}

\theoremstyle{definition}

\theoremstyle{remark}

\allowdisplaybreaks

\usepackage{listings}
\usepackage{xcolor}
\definecolor{codegreen}{rgb}{0,0.6,0}
\definecolor{codegray}{rgb}{0.5,0.5,0.5}
\definecolor{codepurple}{rgb}{0.58,0,0.82}
\definecolor{backcolour}{rgb}{1,1,1}

\lstdefinestyle{mystyle}{
    backgroundcolor=\color{backcolour},   
    commentstyle=\color{codegreen},
    keywordstyle=\color{magenta},
    numberstyle=\tiny\color{codegray},
    stringstyle=\color{codepurple},
    basicstyle=\ttfamily\footnotesize, 
    breakatwhitespace=false,         
    breaklines=true,                 
    captionpos=b,                    
    keepspaces=true,                 
    numbers=left,                    
    numbersep=5pt,                  
    showspaces=false,                
    showstringspaces=false,
    showtabs=false,                  
    tabsize=4,
    frame=lines      
}

\title{Training Graph Foundation Models on The Web Graph}

\author{\name Ryoma Sato \email rsato@nii.ac.jp \\
  \addr National Institute of Informatics
}

\begin{document}

\maketitle

\begin{abstract}
We introduce Acacia, a graph foundation model, trained on the web graph. Acacia (i) supports arbitrary feature dimensionalities and semantics without additional training, (ii) supports a wide range of tasks, including node classification, link prediction, node clustering, and graph generation, without additional training, (iii) has in-context learning capabilities, and (iv) does not rely on pretrained LLMs. In particular, existing graph foundation models often require training additional classification heads or feature projectors to accommodate new graphs or new labels, whereas Acacia does not. Moreover, existing graph foundation models often gain their capabilities by being stitched together with pretrained LLMs, whereas Acacia is trained from scratch using only the Common Crawl web graph. This is also an important result because it provides evidence that graph models can acquire emergent capabilities from scratch like LLMs.
\end{abstract}

\section{Introduction}

Graph structures appear in many forms, including social networks, citation networks, transportation networks, chemical compounds, text (one-dimensional sequences), and images (two-dimensional grids). Conventional approaches have built separate models for each graph domain or task. Models such as graph neural networks can use a single architecture for all kinds of graphs, but a particular set of model weights can only be used with a fixed feature dimensionality, fixed feature semantics, and a fixed task.

Many graph foundation models that can be applied to various tasks have been proposed in recent years. However, many of them depend on the representational power of LLMs or often require additional classification heads or feature projectors to accommodate new graphs or new labels \cite{lachi2025graphfm,yu2025samgpt}. One For All \cite{liu2024one} addresses the mismatch between feature representations across domains by expressing node features as text, feeding them into an LLM, and using the resulting representations as node embeddings. However, much of its capability depends on the LLM, and it is difficult to apply to features that are hard to express as text. AnyGraph \cite{xia2026anygraph} is based on a mixture of experts represented by GCNs and residual MLPs, and improves its capabilities through task-dependent routing, but does not achieve complete alignment for new graphs or new labels.

We propose Acacia\footnote{Acacia is a tree. A tree is a graph. Therefore, Acacia is a graph.}, the first graph foundation model that (i) supports arbitrary feature dimensionalities and semantics without additional training, (ii) supports a wide range of tasks, including node classification, link prediction, node clustering, and graph generation, without additional training, (iii) has in-context learning capabilities, and (iv) does not rely on pretrained LLMs.

One distinguishing feature of Acacia is that it is trained exclusively on the Common Crawl web graph. Modern LLMs have developed through training on web data. Extending this idea to graph foundation models is a natural step. In particular, the web itself has a graph structure, and one could even argue that it is better suited to graph models than to text models. The web's link structure contains diverse topologies that are connected in a ``natural'' way. Learning this structure is expected to enable graph foundation models to handle a variety of graph structures.

Acacia is a decoder-only autoregressive Transformer model, like GPT. Acacia has three types of token, i.e., nodes, edges, and labels, which it generates sequentially. This approach enables a variety of tasks. Generating everything from the beginning yields a graph generation model. Given a graph, filling in its nodes and edges as input to Acacia and generating labels as a continuation enables node classification or node clustering. If the labels of some nodes are known, filling in those labels as well enables transductive classification. Retrieving examples from other labeled graphs and prepending them as separate connected components enables in-context learning. Thus, a single model can perform various tasks by using a decoder-only autoregressive Transformer, as in LLMs, and choosing how to arrange the input tokens.

We trained Acacia-315m, a model with 315M parameters, on 11 billion tokens, obtaining model weights that achieve nontrivial node clustering accuracy in a fully unsupervised setting and possess in-context learning capabilities. These model weights are publicly available on Hugging Face\footnote{\url{https://huggingface.co/joisino/acacia-315m}}. Obtaining a graph foundation model with such general capabilities entirely from scratch, without relying on pretrained LLMs, is an important result in itself.

\section{Acacia}

\begin{figure}[t]
  \centering
  \includegraphics[width=0.65\textwidth]{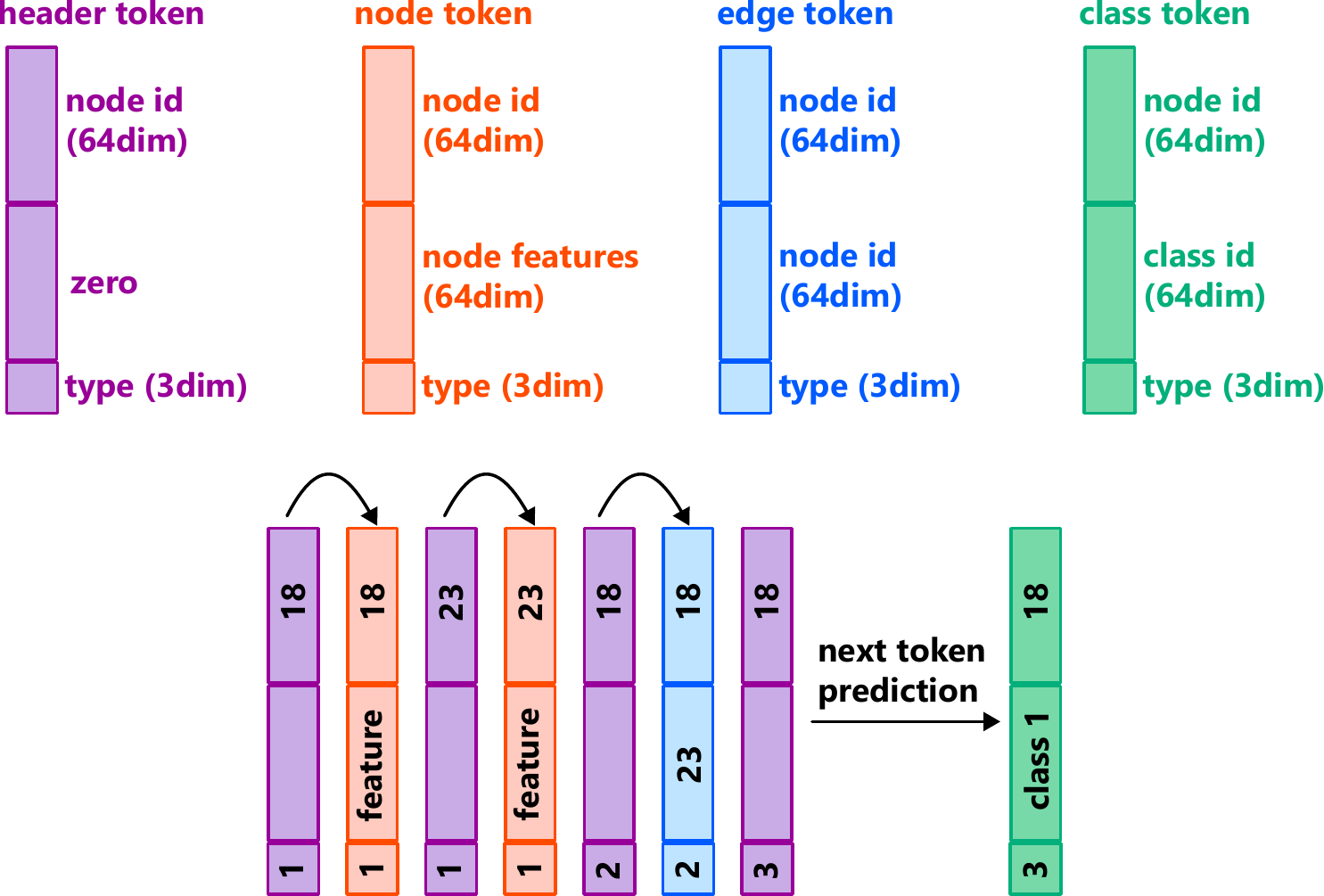}%
  \caption{Token types. Acacia has four types of tokens: headers, nodes, edges, and node labels.} \label{fig: type}
\end{figure}

\subsection{Overview}

Acacia is a decoder-only autoregressive Transformer model, like GPT. Acacia has four types of tokens: headers, nodes, edges, and node labels. Figure \ref{fig: type} shows the vector structure of each token type. In every case, the last 3 dimensions encode the token type as a one-hot vector. A header is a special token that specifies the type of the next token. For a header, the type in the last block refers to the next token rather than to the header itself. The first $(d - 3) / 2$ dimensions represent the node ID, and the following $(d - 3) / 2$ dimensions are always zero. For a node, the first $(d - 3) / 2$ dimensions represent the node ID, and the following $(d - 3) / 2$ dimensions represent the node features. For an edge, the first $(d - 3) / 2$ dimensions and the following $(d - 3) / 2$ dimensions represent the IDs of its endpoints. For a node label, the first $(d - 3) / 2$ dimensions represent the node ID, and the following $(d - 3) / 2$ dimensions represent the label ID. This scheme allows all token types to be treated uniformly.

\begin{figure}[t]
  \centering
  \includegraphics[width=0.65\textwidth]{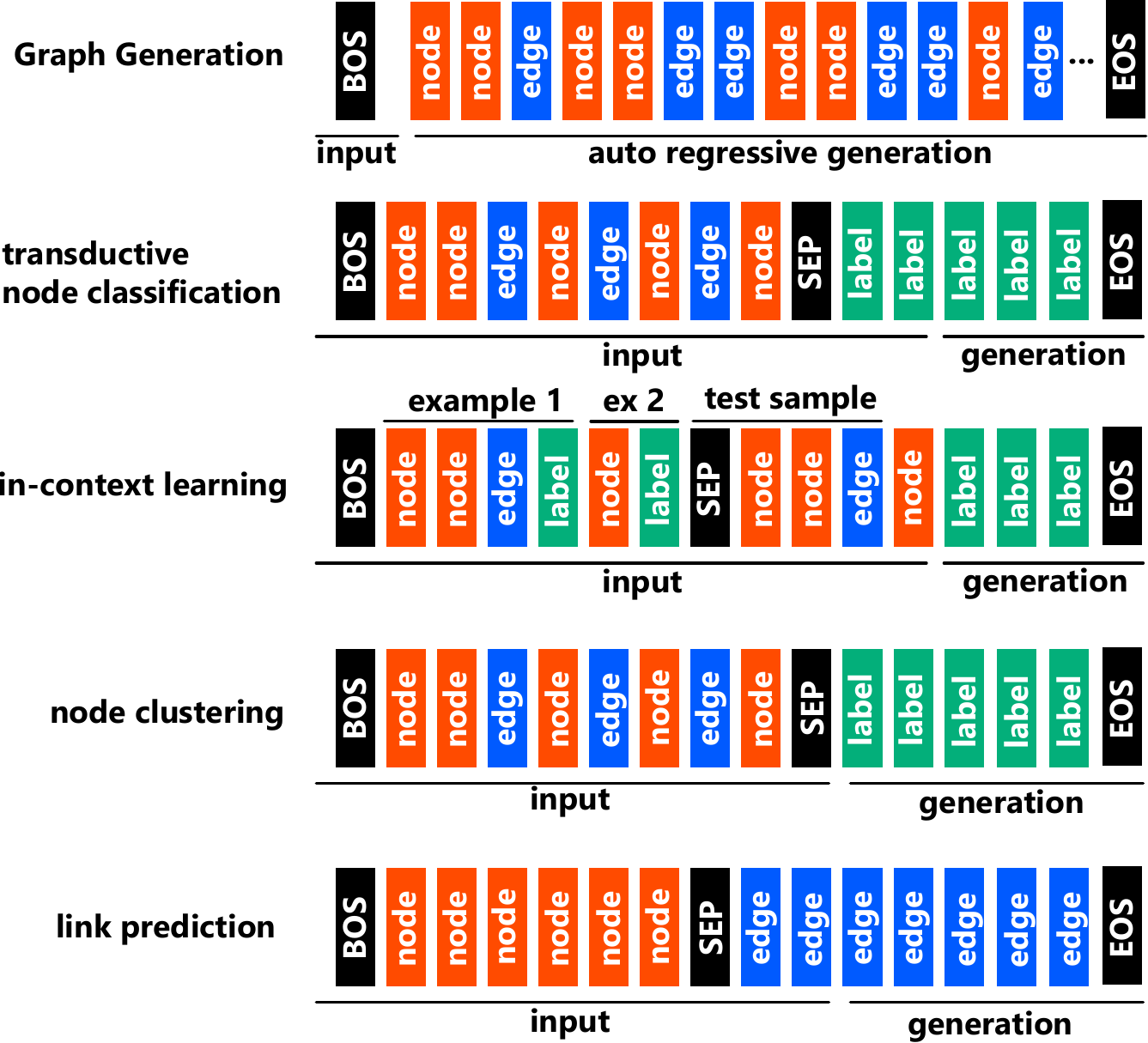}%
  \caption{Examples of tasks supported by Acacia. Different tasks can be handled by changing the tokens included in the prompt and the generation targets. These are only examples; tasks such as jointly predicting edges and node labels are also supported.} \label{fig: tasks}
\end{figure}

Node IDs and label IDs are assigned random vectors that remain consistent within an input. Note that these are not one-hot vectors. This scheme can accommodate arbitrary numbers of nodes and labels. In a high-dimensional space, most vectors are nearly orthogonal and therefore easy to distinguish. Although node IDs carry no meaning, assigning consistent node IDs within a graph allows the model to recognize connectivity. For simplicity, consider identity matrices for both the collection of ID vectors and the embedding matrix (i.e., one-hot node IDs serve directly as embeddings within the Transformer). In an attention layer, when nodes serve as queries and edges as keys, appropriate masking through the projection matrices allows a node to attend strongly only to edges incident to it. For example, let $d = 13$, and let the embeddings of node 3, node 5, and the edge $e$ connecting them be \begin{align}
  \boldv_3 &= [0, 0, 1, 0, 0, 0, 0, 0, 0, 0, 1, 0, 0]^\top \\
  \boldv_5 &= [0, 0, 0, 0, 1, 0, 0, 0, 0, 0, 1, 0, 0]^\top \\
  \boldv_e &= [0, 0, 1, 0, 0, 0, 0, 0, 0, 1, 0, 1, 0]^\top
\end{align} and let the query and key projection matrices be \begin{align}
  \boldW_q &= \begin{pmatrix}
    1 & 0 & 0 & 0 & 0 & 0 & 0 & 0 & 0 & 0 & 0 & 0 & 0 \\
    0 & 1 & 0 & 0 & 0 & 0 & 0 & 0 & 0 & 0 & 0 & 0 & 0 \\
    0 & 0 & 1 & 0 & 0 & 0 & 0 & 0 & 0 & 0 & 0 & 0 & 0 \\
    0 & 0 & 0 & 1 & 0 & 0 & 0 & 0 & 0 & 0 & 0 & 0 & 0 \\
    0 & 0 & 0 & 0 & 1 & 0 & 0 & 0 & 0 & 0 & 0 & 0 & 0 \\
  \end{pmatrix} \\
  \boldW_k &= \begin{pmatrix}
    1 & 0 & 0 & 0 & 0 & 1 & 0 & 0 & 0 & 0 & 0 & 0 & 0 \\
    0 & 1 & 0 & 0 & 0 & 0 & 1 & 0 & 0 & 0 & 0 & 0 & 0 \\
    0 & 0 & 1 & 0 & 0 & 0 & 0 & 1 & 0 & 0 & 0 & 0 & 0 \\
    0 & 0 & 0 & 1 & 0 & 0 & 0 & 0 & 1 & 0 & 0 & 0 & 0 \\
    0 & 0 & 0 & 0 & 1 & 0 & 0 & 0 & 0 & 1 & 0 & 0 & 0 \\
  \end{pmatrix}
\end{align} Then, \begin{align}
  \boldW_q \boldv_3 = [0, 0, 1, 0, 0]^\top \\
  \boldW_q \boldv_5 = [0, 0, 0, 0, 1]^\top \\
  \boldW_k \boldv_e = [0, 0, 1, 0, 1]^\top
\end{align}
Thus, the corresponding inner products are $1$, whereas the inner products between other nodes and edge $e$ are $0$. The same argument holds for non-one-hot IDs as long as they are orthogonal. This encoding scheme is therefore well suited to graph structures. In practice, instead of using such manually constructed weight matrices, the weights are further optimized for the task in a data-driven manner. Moreover, the assignments of node IDs and label IDs are permuted on each run, allowing the model to focus on relationships rather than memorizing particular IDs.

Acacia assumes that header tokens and other tokens alternate in the input. After a header token is provided, the model is trained to generate a token of the type specified in the third block for the node ID specified in the first block. This allows the generation target to be controlled. For example, after providing a token sequence representing a graph, we can predict the label of node $18$ by appending a header token that specifies node $18$ and the label token type, and then performing next-token prediction (bottom of Figure \ref{fig: type}). Similarly, to predict an edge connected to node $18$, we can append a header token that specifies node $18$ and the edge token type, and use next-token prediction to generate one edge connected to that node. In general, users can generate a graph in their desired order by repeatedly adding a header token manually, letting Acacia generate the next token, adding another header token, and letting Acacia generate the next token again. For tasks such as graph generation, Acacia can also generate the header tokens automatically, allowing it to decide which nodes and edges to generate and in what order. In what follows, we omit header tokens from the illustrations when they are obvious.

\begin{figure}[t]
  \centering
  \includegraphics[width=0.65\textwidth]{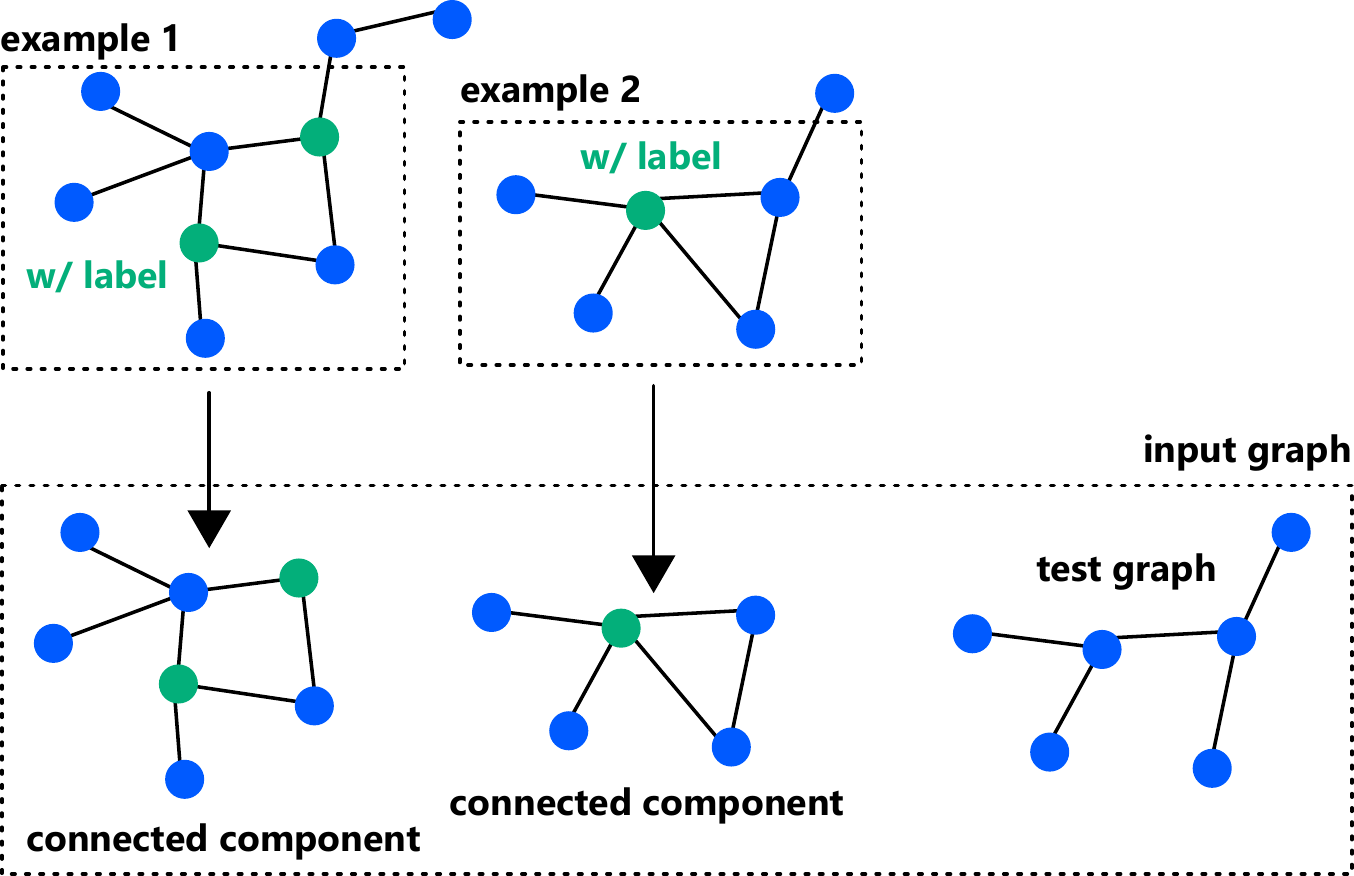}%
  \caption{An example of in-context learning. In-context learning is naturally enabled by taking entire labeled graphs or extracting ego graphs around labeled nodes and adding them to the test graph as new connected components. The example graphs can be fixed, or graphs with structures similar to the test graph can be retrieved from a database and used as examples, as in retrieval-augmented generation (RAG).} \label{fig: icl}
\end{figure}

Combining these tokens enables various tasks (Figure \ref{fig: tasks}). Because Acacia is a decoder-only autoregressive Transformer generative model, it generates a graph when no conditioning information is provided. Given a graph, we can prefill the prompt with that graph and generate a continuation to predict the remaining labels or edges. In particular, it naturally supports transductive node classification, in which labels are given for some nodes and the labels of the remaining nodes are predicted. If no labels are provided in the input, the task becomes unsupervised clustering, with no predefined binding for the labels. With some exceptions \cite{pakman2020attentive}, conventional clustering methods require the number of classes to be specified in advance; another advantage of this approach is that it does not. Furthermore, as illustrated in Figure \ref{fig: icl}, in-context learning is naturally enabled by taking entire labeled graphs or extracting ego graphs around labeled nodes and adding them to the test graph as new connected components. The example graphs can be fixed, or graphs with structures similar to the test graph can be retrieved from a database and used as examples, as in retrieval-augmented generation (RAG) \cite{lewis2020retrieval}.

\subsection{Training on the Web Graph}

\begin{table}[t]
\centering
\caption{Sampling probabilities for the elements provided as conditions and those predicted autoregressively.}
\label{tab: autoregressive_sampling}
\begin{tabular}{lll}
\hline
Conditioning elements & Autoregressive targets & Sampling probability \\
\hline
nodes + edges  & labels                  & 50\% \\
nodes          & edges + labels          & 10\% \\
nodes + labels & edges                   & 10\% \\
labels         & nodes + edges           & 7.5\% \\
edges          & nodes + labels          & 7.5\% \\
edges + labels & nodes                   & 7.5\% \\
None           & nodes + edges + labels  & 7.5\% \\
\hline
\end{tabular}
\end{table}

We train Acacia on the Common Crawl web graph. Nodes represent web pages, edges represent hyperlinks, and labels represent domains. Training graphs are extracted from the Common Crawl web graph using breadth-first search starting from 1 or more randomly selected nodes. The model predicts domains from link structures or additional links from the connections between web pages. To support different prediction types, we mix different ways of providing token types, as shown in Table \ref{tab: autoregressive_sampling}. Tokens are randomly shuffled within the conditioning part and within the autoregressive prediction part separately. The number of nodes $n$ is sampled from $8$ to $128$ according to $P(n) \propto n^{-1.1}$, and breadth-first search proceeds until that number of nodes is reached. This enables the model to handle graphs of various sizes. Training only on large graphs can be too difficult for learning to progress, whereas including graphs with fewer nodes allows learning to progress even in the early stages. We expect this implicit curriculum to facilitate the smooth acquisition of capabilities.

In addition to these tasks, we also construct some graphs as shown in Figure \ref{fig: icl} to develop in-context learning capabilities.

\subsection{A Structure Independent of Feature-Dimension Alignment}

One difficulty in building graph foundation models is that the number and meaning of feature dimensions vary across graphs. In citation network data, dimension $1$ might indicate whether a paper contains the word ``language,'' whereas in chemical compound data, dimension $1$ might indicate whether an atom is carbon. The input vector alone does not reveal which meaning applies. A model trained on the former data can therefore be confused when presented with the latter data at test time.

Acacia deliberately corrupts the information in training features to reduce its dependence on individual dimensions. Specifically, node features are constructed using signed hashing of the text in web page titles and bodies. For example, each occurrence of ``language'' might add $+1$ to dimension $19$, while each occurrence of ``tour'' might add $-1$ to dimension $53$. This converts the text into a $64$-dimensional hashed bag of words. As a result, each dimension mixes multiple meanings instead of having a specific, concrete meaning.

To further encourage a focus on relationships, we also replace the features of some randomly selected nodes with samples from the standard normal distribution. In particular, with a certain probability, we create examples in which the features of every node in the graph are replaced with samples from the standard normal distribution. This is inspired by theoretical results showing that random features enhance the capabilities of graph neural networks, enabling them to solve various combinatorial problems on graphs \cite{sato2020random, sato2019approximation, sato2020survey}. This is also useful in practice because it enables the model to handle cases where node features are unavailable at test time.

At test time, we project node features into $64$ dimensions using a random projection matrix. This matrix is shared within each graph and corresponds to the hashing used during training. This allows the model to be applied to graphs with different feature dimensionalities and semantics, while still processing the graphs using relationships among dimensions and how those dimensions relate to one another within the graph structure.

\subsection{The Acacia Recipe}

We describe the specific recipe for Acacia-315M. Acacia's architecture is inspired by SmolLM2-360M \cite{allal2025smollm2}. Specifically, it is a decoder-only model with 32 layers, a hidden dimension of 960, and 315M parameters. However, we draw only on its architectural configuration and do not use any pretrained weights.

We first trained the model for 10 billion tokens in a simple setting, followed by an additional 1 billion tokens of training. In the simple setting, only web page titles are used to construct node features, and no data are explicitly constructed for in-context learning. During the additional training, we use the first 512 words of the page body, which is more informative than the title, to construct node features, and mix in graphs explicitly constructed for in-context learning as shown in Figure \ref{fig: icl}. Note that our 10 billion tokens are not directly comparable to 10 billion tokens in LLM training. In LLM training, approximately one word corresponds to one token. In our training, by contrast, a single web page corresponds to one to a few tokens. Since a web page contains hundreds of tokens, Acacia may process hundreds of times as many web pages even at the same total of 10 billion tokens.

Training starts entirely from scratch with random initialization. We use AdamW, warming up the learning rate from 3e-6 to 3e-4 over 20M tokens and then decaying it to 3e-5 over the remaining tokens with a cosine schedule. For the additional training, we reinitialize AdamW, warm up the learning rate from 3e-6 to 1e-4 over 10M tokens, and then decay it to 1e-5 with a cosine schedule.

Initial training used 8 RTX 5090 GPUs and completed in 20 hours. The additional training used 4 RTX 5090 GPUs and 4 L40S GPUs and completed in 4 hours. Training took approximately 24 hours in total.

\section{Experiments}

\begin{table}[t]
    \centering
    \caption{Dataset statistics and official data splits.}
    \label{tab: dataset_statistics}
    \begin{tabular}{lrrrr}
        \hline
        Dataset & Total nodes & Classes & Original feature dim. & Official train / val / test \\
        \hline
        Cora          & 2,708     & 7  & 1,433 & 140 / 500 / 1,000 \\
        CiteSeer      & 3,327     & 6  & 3,703 & 120 / 500 / 1,000 \\
        PubMed        & 19,717    & 3  & 500   & 60 / 500 / 1,000 \\
        ogbn-products & 2,449,029 & 47 & 100   & 196,615 / 39,323 / 2,213,091 \\
        \hline
    \end{tabular}
\end{table}

We experimentally evaluate the performance of Acacia-315M. In particular, we examine whether the foundation model works without additional training. Acacia-315M can adapt to various tasks and datasets without updating its model weights.

We use Cora, CiteSeer, PubMed, and ogbn-products as standard testbeds. We use the Planetoid public splits \cite{sen2008collective} for Cora, CiteSeer, and PubMed, and the official OGB split \cite{hu2020open} for ogbn-products. Table \ref{tab: dataset_statistics} lists their statistics. These datasets differ in feature dimensionality and labels, and naturally also in the meanings of their feature dimensions and labels. In particular, their features, labels, and graph structural properties differ from those of the web graph used for pretraining. Nevertheless, Acacia achieves nontrivial accuracy on these datasets.

\subsection{Node Classification}

\begin{table}[t]
    \centering
    \caption{Node classification performance. Each value is the mean accuracy $\pm$ sample standard deviation over 3 seeds.}
    \label{tab: node_classification}
    \begin{tabular}{lcccc}
        \toprule
        & Cora & CiteSeer & PubMed & ogbn-products \\
        \midrule
        Chance level
        & $41.94 \pm 0.29$
        & $32.38 \pm 0.13$
        & $19.68 \pm 0.72$
        & $44.15 \pm 0.66$ \\
        Untrained, frozen
        & $41.07 \pm 1.37$
        & $33.07 \pm 0.67$
        & $19.80 \pm 0.56$
        & $45.00 \pm 0.17$ \\
        Acacia, frozen
        & $58.17 \pm 1.35$
        & $39.50 \pm 1.22$
        & $19.87 \pm 0.74$
        & $66.00 \pm 0.72$ \\
        \bottomrule
    \end{tabular}
\end{table}

We first perform standard node classification. For each test node, we extract a neighborhood graph of up to 128 nodes using breadth-first search, assign labels only to nodes with training labels, and tokenize the graph. We then predict the label of the test node. By its nature, Acacia can ground labels that appear in the input graph, but cannot ground labels that do not; it can only recognize them as ``new labels.'' In this experiment, we therefore restrict the output to labels present in the input graph. If no node in the neighborhood has the same label as the test node's true label, the prediction is necessarily counted as incorrect. Conversely, if the input graph contains only nodes labeled with the test node's true label and unlabeled nodes, even the untrained model is necessarily counted as correct. This occurs frequently in Cora, Citeseer, and ogbn-products because of homophily, making the chance level higher than that of uniform selection from all labels. In contrast, labels are sparse in PubMed, so the entire neighborhood is often unlabeled. Predictions are necessarily counted as incorrect in such cases, making the chance level lower than that of uniform selection from all labels. Note that the weights of Acacia-315M are never updated in this experiment.

Table \ref{tab: node_classification} presents the results. On the 3 datasets other than PubMed, Acacia outperforms both the randomly initialized, untrained Acacia model and the chance level. PubMed is the exception because only 60 of its 19,717 nodes are labeled: in many cases, the extracted graph contains no training labels or no node with the correct label, so the prediction is necessarily counted as incorrect. In-context learning, described next, is useful in such cases.

\subsection{In-Context Learning}

\begin{table}[t]
    \centering
    \caption{Classification performance with In-Context Learning. Each value is the mean accuracy $\pm$ sample standard deviation over 3 seeds.}
    \label{tab: in_context_learning}
    \begin{tabular}{lcccc}
        \toprule
        & Cora & CiteSeer & PubMed & ogbn-products \\
        \midrule
        Chance level
        & $14.29$
        & $16.67$
        & $33.33$
        & $2.35$ \\
        Untrained, frozen
        & $15.27 \pm 0.92$
        & $17.30 \pm 1.57$
        & $33.67 \pm 1.03$
        & $2.37 \pm 0.40$ \\
        Acacia, frozen
        & $54.33 \pm 1.61$
        & $39.37 \pm 2.06$
        & $37.43 \pm 2.05$
        & $68.38 \pm 0.32$ \\
        \bottomrule
    \end{tabular}
\end{table}

For each test node, we extract a neighborhood graph of up to 32 nodes using breadth-first search. We additionally select 1 to 2 examples per class from the training nodes and place them in connected components separate from the test node. There are no edges between components, and no nodes are shared between them. The entire graph contains at most 128 nodes. Because labels are always present in the input, the model can select one of them, almost eliminating cases in which a prediction is necessarily incorrect. We say ``almost'' because ogbn-products has labels that appear in the test data but not in the training data. In such cases, the true label is absent from the examples, so the prediction is necessarily incorrect. This does not occur in Cora, CiteSeer, or PubMed. Note that the weights of Acacia-315M are never updated in this experiment.

Table \ref{tab: in_context_learning} presents the results. On every dataset, Acacia outperforms both the randomly initialized, untrained Acacia model and the chance level. In particular, it exceeds the chance level even on PubMed, where performance without in-context learning was close to chance. This is an important result because it demonstrates that LLM-style in-context learning through token prompts is possible using graph data alone, without relying on pretrained LLMs.

\subsection{Clustering}

\begin{table}[t]
    \centering
    \caption{Clustering performance of the untrained model and Acacia. Each value is the mean $\pm$ sample standard deviation over 3 seeds.}
    \label{tab: acacia_clustering_results}
    \begin{tabular}{lcccc}
        \toprule
        & Cora & CiteSeer & PubMed & ogbn-products \\
        \midrule
        Untrained (4 nodes, ARI)
        & $0.0013 \pm 0.0039$
        & $0.0008 \pm 0.0036$
        & $0.0009 \pm 0.0030$
        & $0.0016 \pm 0.0015$ \\
        Acacia (4 nodes, ARI)
        & $0.2643 \pm 0.0120$
        & $0.2833 \pm 0.0311$
        & $0.2034 \pm 0.0043$
        & $0.0117 \pm 0.0069$ \\
        \midrule
        Untrained (4 nodes, AMI)
        & $0.0006 \pm 0.0024$
        & $0.0025 \pm 0.0046$
        & $0.0003 \pm 0.0018$
        & $0.0006 \pm 0.0006$ \\
        Acacia (4 nodes, AMI)
        & $0.3295 \pm 0.0053$
        & $0.3241 \pm 0.0178$
        & $0.2627 \pm 0.0121$
        & $0.0222 \pm 0.0106$ \\
        \midrule
        Untrained (1 node, ARI)
        & $-0.0013 \pm 0.0029$
        & $0.0050 \pm 0.0167$
        & $0.0006 \pm 0.0047$
        & $0.0009 \pm 0.0023$ \\
        Acacia (1 node, ARI)
        & $0.1191 \pm 0.0084$
        & $0.0667 \pm 0.0304$
        & $0.0387 \pm 0.0068$
        & $0.0047 \pm 0.0116$ \\
        \midrule
        Untrained (1 node, AMI)
        & $-0.0009 \pm 0.0046$
        & $0.0096 \pm 0.0116$
        & $0.0039 \pm 0.0011$
        & $0.0001 \pm 0.0039$ \\
        Acacia (1 node, AMI)
        & $0.1437 \pm 0.0150$
        & $0.0792 \pm 0.0341$
        & $0.0599 \pm 0.0035$
        & $0.0061 \pm 0.0099$ \\
        \bottomrule
    \end{tabular}
\end{table}

Next, we conduct challenging clustering experiments without using any labels. We construct input graphs in two ways. The first randomly selects 4 nodes, extracts a graph by breadth-first search from each, and uses their union as the input graph. These generally form a disconnected graph, making the clustering structure relatively clear. The second randomly selects 1 node and extracts a graph using breadth-first search. This produces a contiguous region of the graph, making clustering more difficult. Given a graph, we provide only its nodes and edges to Acacia and generate labels sequentially. Note that the weights of Acacia-315M are never updated in this experiment.

Table \ref{tab: acacia_clustering_results} reports the results. We use the Adjusted Rand Index (ARI) and Adjusted Mutual Information (AMI) as clustering metrics. The true node labels are used only for evaluation. The chance level is $0$ for both metrics. In every setting and on every dataset, Acacia outperforms both the randomly initialized, untrained Acacia model and the chance level. This shows that Acacia can effectively process graph structures even when no labels are used.

\section{Related Work}

There have been many attempts to build graph foundation models \cite{liu2025graph,mao2024position,tang2025toward,shen2025zero,xia2026anygraph,wang2025towards,chen2026towards,zhao2025graphany,chen2025flatten}. However, to the best of our knowledge, none pretrain on the web graph. Instead, common approaches pretrain on existing graph datasets such as citation or molecular graphs \cite{he2025unigraph2,wang2025multi,zhao2025graphgpt,chen2025graph,yuan2026rag,ma2025gilt}, on in-house data \cite{speicher2026billion,he2025generalizing}, or on synthetic data \cite{xia2024opengraph,eremeev2025graphpfn,platonov2026fair,choi2026learning}. These approaches may improve downstream performance on related datasets, but do not necessarily guarantee generality to other tasks. Instead, we aim to acquire general capabilities without excessive manual design of inductive biases by avoiding overly curated data and using the Common Crawl web graph as a neutral source of data.

Although we perform in-context learning, methods for in-context learning on graph data already exist \cite{huang2023prodigy,guan2026advancing,zhuo2026modality}. However, they may use GNNs, as in PRODIGY \cite{huang2023prodigy}, or have difficulty generalizing to unseen graphs with different node features, as in VISION \cite{guan2026advancing}. Our strength is that our model can be applied to new graphs with different node features and that we build GPT-style in-context learning from scratch.

Recent advances in LLMs have also prompted efforts to develop graph foundation models based on LLMs \cite{he2025unigraph,kong2025gofa,kimino2026why,sun2025graphicl,li2025are}. Similarly, advances in tabular foundation models have motivated graph foundation models based on models such as TabPFN \cite{hayler2025of,eremeev2025turning,choi2025can,liao2026tfmlinker}. Because these approaches build on established large models, they tend to achieve high accuracy, but they are difficult to apply to data that are hard to convert into text or tables. From a scientific perspective, it is also difficult to distinguish whether their performance comes from the capabilities of the pretrained model or from graph learning. Our position is that, although these directions should be pursued for practical purposes, advancing graph learning also requires research on training pure graph models without relying too heavily on models from other domains. We believe our work advances graph learning by training a pure graph model from scratch, without relying on pretrained LLMs or similar models, and acquiring emergent capabilities such as in-context learning and zero-shot learning.

\section{Conclusion}

We proposed Acacia and trained it on the web graph. Acacia is the first graph foundation model that (i) supports arbitrary feature dimensionalities and semantics without additional training, (ii) supports a wide range of tasks, including node classification, link prediction, node clustering, and graph generation, without additional training, (iii) has in-context learning capabilities, and (iv) does not rely on pretrained LLMs. Acacia learns diverse graph structures from the web graph and achieves nontrivial downstream performance without adaptation. This is also an important result because it provides evidence that graph models can acquire emergent capabilities from scratch like LLMs.

\bibliography{main}
\bibliographystyle{abbrvnat}

\appendix

\end{document}